\documentclass[conference]{IEEEtran}
\IEEEoverridecommandlockouts
\usepackage{cite}
\usepackage{float}
\usepackage{amsmath,amssymb,amsfonts}
\usepackage[ruled,vlined]{algorithm2e}
\usepackage{graphicx}
\usepackage{gensymb}
\usepackage{textcomp}
\usepackage{xcolor}
\usepackage{steinmetz}
\usepackage{balance}
\def\BibTeX{{\rm B\kern-.05em{\sc i\kern-.025em b}\kern-.08em
    T\kern-.1667em\lower.7ex\hbox{E}\kern-.125emX}}
\begin{document}

\title{A Neuromorphic Model of Learning Meaningful Sequences
with Long-Term Memory}

\author{\IEEEauthorblockN{Laxmi R. Iyer}
\IEEEauthorblockA{\textit{Dept. of Applied Informatics}\\
\textit{Comenius University}\\
Bratislava, Slovakia \\
Laxmi.Iyer@fmph.uniba.sk}
\and
\IEEEauthorblockN{Ali A. Minai}
\IEEEauthorblockA{\textit{Dept. of Electrical} \\
\textit{and Computer Engineering}\\
\textit{University of Cincinnati}\\
Cincinnati, USA \\
minaiaa@ucmail.uc.edu}
}

\maketitle

\begin{abstract}
Learning meaningful sentences is different from learning a random set of words. When humans understand the meaning, the learning occurs relatively quickly. 
What mechanisms enable this to happen? In this paper, we examine the learning of novel sequences in familiar situations. We embed the Small World of Words (SWOW-EN), a Word Association Norms (WAN) dataset, in a spiking neural network based on the Hierarchical Temporal Memory (HTM) model to simulate long-term memory. Results show that in the presence of SWOW-EN, there is a clear difference in speed between the learning of meaningful sentences and random noise. For example, short poems are learned much faster than sequences of random words. In addition, the system initialized with SWOW-EN weights shows greater tolerance to noise. 
\end{abstract} 
 
\section{Introduction}
\label{introduction} 
Humans have a deeply grounded understanding of words in a language. Concepts are learned through experience and associated with words. Further, humans are able to associate them with related concepts to put together sentences, analogies, stories, and so on. They are then able to convey information, and communicate facts, thoughts, and even feelings effectively with others. Ultimately, this ability must arise from a biological substrate of neurons and synapses, and biological principles such as connectivity, distributed representations, and synaptic activation. 

In this paper, we address a specific question on meaning. When we learn something new, such as a new song or poem, the learning does not happen from scratch. Although the song or poem being learned might be new, we know the language and understand the meaning of the poem. We may have some additional information about the domain. For example, we know what to expect on a farm, or in the sky at night. This web of associations facilitates our learning process and enables us to learn quickly. If, for example, we learn a song in a language we do not know, we may actually be able to learn the lyrics, but it will take much longer to do so. In this paper, we want to model and simulate how prior semantic associations can accelerate the learning of meaningful sequences compared to random ones. 

The paper is organized as follows. In the next section we present the background work. In the following section, we introduce the problem of learning novel sequences in familiar situations. We then present the system architecture which contains both a brief description of the HTM architecture on which our system is based, and the features we have added to the system. We present our simulations and results, which is followed by a discussion and conclusion. 

\section{Background - The Search for Meaning}
\label{background}

Neurobiological findings show that semantic information is represented in the brain at different levels in terms of concepts, words, and features that ground the concepts to sensorimotor affordances, and contexts that highlight different nuances of the word in different situations.  Sensorimotor features are processed in lower-level association cortices and premotor regions. For example, while reading action words, the motor cortex is activated \cite{Hauk:2011, Carota:2017}, passive reading of taste related words activates the gustatory cortices \cite{Barros:2011}, and examining the modality-specific properties of words activates sensory regions in the brain \cite{Barsalou:1999,Barsalou:2003,Barsalou:2008}. This phenomenon provides a basis for understanding language acquisition. A computational neurobiologically constrained model of this has been implemented by Tomasello et al. \cite{Tomasello:2018}. In this model, 12 cortical areas that include primary, secondary, and higher association areas have been modeled. The model includes articulatory, auditory, sensory and motor areas. It demonstrates learning of semantic information from experience, and explains the existence of both modal (category-specific) and amodal representations of concepts. 

Multimodal representations of concepts are stored in higher order association regions. Semantic memory is guided and controlled, in the presence of various contexts, by the lateral prefrontal cortex (PFC), the dorsolateral prefrontal cortex (DLPFC), the dorsomedial prefrontal cortex, and the posterior middle temporal gyrus \cite{Whitney:2011}. The basal ganglia functions as the gate for the internal attention function of cognitive control.

Earlier, our group had implemented a connectionist model of idea generation called CANDID {\cite{Iyer:2009,Iyer:2021,Minai:2021a,Iyer:2010,Minai:2021b}, 
based on both neurobiological findings and behavioral experiments. This model demonstrated that ideas arise from the interaction of semantic categories, concepts and sensorimotor and higher-level features in the presence of a context. The interaction of semantic information at various levels has also been modeled mathematically by Polyn et. al. \cite{Polyn:2009}. Based on a large number of behavioral studies on memory, they developed the Context Memory and Retrieval (CMR) model \cite{Polyn:2009}. They describe memory search as a multi-constrained iterative process, that is shaped by associative connections with other concepts, influences from features, and semantic source information. The memory search could result in further changes in context, as the influences are reciprocal. Both our model
and CMR adopt the idea of ``spotlights for memory'' – that context representations and semantic categories play the role of an attentional spotlight and highlight the items in memory that are active and will play a part in semantic function.  

Drawing upon both models, we present the {\it Multi-Constrained Memory} (MCM) model, which is based on two postulates:

\begin{enumerate}
\item Semantic information is represented at different levels, namely concepts, words, categories and features.

\item Learning and search in semantic memory are constrained and iterative processes.  They are modulated by many different influences, predominantly, context, concepts which are associativity linked to one another, and features.
\end{enumerate}

MCM is derived from principles that are similar to the CANDID model. It does not restrict itself to a particular task, such as ideation, but is a generic model for semantic memory. 

Rather than modeling all the different kinds of semantic representations, we focus in this paper only on word-word associations using Word Association Norms (WANs). Psychologists have built databases of WANs \cite{Nelson:2004,Deyne:2019} from behavioral studies. Subjects are given a cue word and asked the first word or first few words that come to mind. Directed graphs are then built where the edges represent the probabilities that the target word was primed by the cue. Word association norms have been considered by psychologists to be among the best methods for understanding how people associate words in their own minds.  WANs are highly successful in predicting cued recall. Networks built from WANs are specifically suited to model salient psychological phenomena such as developmental shifts \cite{Otgaar:2012}, individual differences in creativity \cite{Gough:1976,Benedek:2012,Kenett:2019}, flexibility of thought \cite{Kenett:2018}, or clinical states such as schizophrenia \cite{Ronald:1964, Brendan:2005, Manschreck:2012}.  

In this paper, we use WANs to model word-word associations, and study the learning of novel sequences constrained by prior semantic associations. 

\section{Learning Novel Sequences in Familiar Situations}
\label{novelLearning}

While human learning is a complex phenomenon, it is known that humans find it much easier to learn in familiar situations. When we have domain knowledge about a particular field, learning occurs much faster. For example, when a language and meaning is known, learning poems occurs much more easily, and an experienced programmer will find it much easier to learn a new programming language. This has been noted in animals as well – for example, animals tend to stay in familiar territories as they find it easier to navigate and learn new routes in familiar territory \cite{Stamps:1995}. However, the mechanisms that facilitate learning in familiar situations are less clear. 

In this paper, we explore the learning of new sequences in the presence of long-term memory. According to \cite{Camina:2017}, ``Long Term Memory refers to unlimited storage of information to be maintained for long periods, even for life''. Long-term memory is divided into declarative and implicit, and declarative memory is further categorized into semantic and episodic memory. In the model, long-term memory is seen as being instantiated by the WAN network called {\it Small World of Words, English} (SWOW-EN) \cite{Deyne:2019}. As we have seen, WANs model how words are associated in the brain. While long-term memory in general consists of far more than these associations, such word associations are certainly an important part of long-term memory in the context of language. In the rest of the paper, we refer to the WAN weights embedded in the system as {\it LTM weights}. 

For sequence learning, we use the Hierarchical Temporal Memory (HTM) \cite{Cui:2016,Hawkins:2016,Hawkins:2017} a biologically constrained computational model. In addition, we use several mechanisms that have been widely observed in neuroscience and psychology experiments, but seldom used in machine learning, including the following:

\begin{enumerate} \label{addedFeas_bkg}
\item \textbf{Gating of Plasticity:} In our system, when a next item in a sequence is expected then learning occurs at a higher rate, else it occurs at a lower learning rate. This allows quick learning of familiar items, and reduces noise. Such a synaptic gating mechanism, triggered by long-term memory, has been observed in the hippocampus \cite{Basu:2016}, where long-range inhibitory projections (LRIPs) to the CA1 region act as a temporally precise disinhibitory gate. Plasticity is induced when cortical input relaying sensory information and hippocampal input conveying information from long-term memory
 arrive at a precise time interval. In our system we
 use an abstraction of such a mechanism: plasticity is increased when information activated by the sensory input matches predictions from long-term memory.
 
\item	\textbf{Rehearsal:} Repeated rehearsals and practice at regular intervals are necessary for the consolidation of long-term memory \cite{Lehmann:2015,Parle:2006,Bonstrup:2019}. Rehearsal involves iterative recall of the learned item \cite{Lehmann:2015}. After repeated rehearsal of one item, the learner is subject to fatigue and needs to switch to another item or rest \cite{Sievers:1944,Ackermann:2017,Bonstrup:2019}. 

\item	\textbf{Weight Decay:} It has been observed for e.g. \cite{Zucker:2002,Leimer:2019,Panda:2018,Tetzlaff:2013} that unless synaptic information is consolidated to long-term memory, it decays over time. While we do not address memory consolidation mechanisms in this paper, we do apply synaptic decay mechanisms. 
\end{enumerate}

While the mechanisms of learning are still a very simplified version of how humans learn, they are more realistic than general deep learning models, where items are repeated continuously and learned through gradient descent. 

\section{System Architecture}
\label{systemArchitecture}


\subsection{HTM} \label{htm}

Hierarchical Temporal Memory (HTM) is a theory of memory and computational system created by Hawkins \cite{Hawkins:2016, Hawkins:2017, Cui:2016}, based on biological principles. The HTM model has a columnar architecture, mimicking the neocortex. It consists of many columns, with each column having several cells. In \cite{Hawkins:2016}, sequence learning is implemented with HTMs. 

In the neuroscience literature \cite{Hawkins:2016}, it is well-known that neurons calculate not just using the soma but also dendrites, which are active and play different roles in computation. Unlike the point neuron used in neural networks, where each neuron has a single weighed sum of all synapses, the HTM incorporates dendritic computation. Various dendritic functions are postulated as follows \cite{Hawkins:2016}: 

\begin{itemize}
	\item Proximal dendrites represent the receptive field of the neuron. They provide input into the neuron. 
	\item Basal dendrites learn the transitions between activity patterns. Such a dendrite depolarizes the soma when a pattern is recognized, indicating a matched prediction. However, it is not enough to fire an action potential. It just causes a neuron to fire earlier than it would have, thus inhibiting neighboring neurons. Thus, it creates a sparse pattern of activity for correctly predicted input. Basal dendrites are used to connect the neuron laterally to other neurons. 
	\item Apical dendrites provide context information.  
\end{itemize}

Both $basal$ and $apical$ dendrites are termed as $distal$ dendrites as they are further away from the soma. In this paper, we do not utilize the $apical$ dendrites, so for the rest of the paper, $basal$ dendrites will be termed as $distal$ dendrites. 

We use the HTM as the basis of our computational model, but add several features as described in the next section and  Section \ref{addedFeas_bkg}.  

\section{System Description}
\label{systemDesc}

We refer to a pattern, $p_k$ as an {\it item} in memory, and a sequence as an {\it ordered se}t of patterns, $s = \{p_1, p_2, … , p_q\}$. A pair of previous and current items in the sequence are known as the $p-c$ pair. The purpose of the system is to learn a set of $m$ sequences $\{s_1, s_2, …, s_m\}$. 

The HTM system consists of $M$ columns mimicking the columns of the cortex. Each column has $N$ cells. As the original HTM system does not use spiking neurons, the authors prefer the term \emph{cell}, and since we are using spiking neurons, we will use the term \emph{neurons}. A set of $b$ columns, $\{c_1, c_2, .., c_b\}$ represent an item in memory. Every neuron $n$ in a column $c$ represents the same information, but in different contexts (or sequences in this paper). Every column receives input from the proximal dendrites. They provide the same input to the entire column. Every neuron is connected to neurons in other columns via the distal dendrites, and the connections are as follows. Each neuron has a set of distal segments, $G_{ij}$, such that $G^d_{ij}$ represents the $d$'th segment of the $i$'th cell in the $j$'th column.  $NM$ is the total number of neurons in the layer. Each distal segment contains a number of synapses which represent lateral connections from a subset of $NM-1$ neurons. $A^t$ is the M $\times$ N binary matrix where $a_{ij}^t$ is the activity of the $i$-th neuron in the $j$-th column at time $t$. $\Pi^t$ is the M $\times$ N binary matrix which denotes neurons in a predictive state, where $\pi^t_{ij}$ is the prediction of the of the $i$-th neuron in the $j$-th column at time $t$. The column is divided into mini-columns, and neurons in a mini-column share the same feedforward receptive fields. The feedforward input activates a set of $k$ columns. Let this set be $Y^t$. The active state for each neuron is calculated as follows: 

\begin{equation}
\label{eqn:firstHTMeqn}
    a^t_{ij}=
    \begin{cases}
      1, & \text{if $j \in Y^t$ and $\pi_{ij}^{t-1}$ }\ = 1, \\
      1, & \text{if $j \in Y^t$ and $\Sigma_i \pi^{t-1}_{ij}$}\ = 0, \\ 
      0, & \text{otherwise} 
    \end{cases}
\end{equation}

If any neuron in a column is predicted, it fires before others in the column, and immediately inhibits all others. Therefore, if there are predicted neurons in the column, only it will be active, and the others will remain inactive, as shown in the first line of the equation. If there are no predicted neurons, then all neurons in the column will be active, as stated in the second line of the equation. 

Each synapse is associated with a {\it permanence}, which is a scalar between $[0, 1]$, and a {\it weight} which is binary, $0$ or $1$. When the permanence crosses a threshold, the weight becomes $1$, else it is $0$. A permanence value at threshold represents a synapse that is not fully formed, and permanence that is close to 1 represents a synapse that is fully formed. A connected synapse indicates a synapse that has crossed the permanence threshold. We use $\widetilde{G}^d_{ij}$ to denote a binary matrix which contains only the connected synapses. A synapse is activated by an active presynaptic neuron. A segment is active when more than $m$ synapses are active. In order for a postsynaptic neuron to be predicted, at least one of its segments should be active. 

The predictive state for the current time step is calculated as follows: 

\begin{equation}
\pi^t_{ij} = 
\begin{cases}
1, & \text{if $\exists_d \left\lVert\widetilde{G}_{ij}^d \circ A^t\right\rVert_1 > \theta$} \\
0, & \text{otherwise}
\end{cases}
\end{equation}

where the threshold $\theta$ represents the spiking threshold and $\circ$ represents the element-wise multiplication. If there are more than $\theta$ connected synapses with active presynaptic neurons, then the segment will be active and generate a spike. A neuron will be depolarized if at least one segment is active. 

We choose those segments $G^d_{ij}$ such that: 

\begin{equation}
\forall_{j \in Y^t} (\pi^{t-1}_{ij} > 0) \; \text{and} \; \left\lVert\widetilde{G}_{ij}^d \circ A^{t-1}\right\rVert_1 > \theta   \\
\end{equation}

The first term selects winning columns that contained the correct predictions, and the second term specifically selects the segments that are responsible for the prediction. Let $\dot{G}^d_{ij}$ denote the binary matrix containing only positive entries in $G^d_{ij}$. We reinforce a segment where the following is true. 

\begin{align}
\forall_{j \in Y^t} (\Sigma_i \pi^{t-1}_{ij} = 0) \; \text{and} \\ \; \left\lVert\dot{G}_{ij}^{*d} \circ A^{t-1}\right\rVert_1 = 
max_i(\left\lVert\dot{G}_{ij}^d \circ A^{t-1}\right\rVert_1) \nonumber
\end{align}

We reward synapses in the following manner. All permanence values are decreased by a small value, $p^-$, and the permanence values of the active presynaptic neurons are increased by a larger value, $p^+_m$.   

\begin{equation}
\label{eqn:learningEqn}
   \triangle G^d_{ij} = p^+_m(\dot{G}^d_{ij} \circ A^{t-1}) - p^-\dot{G}_{ij}^d
\end{equation}

$p^+_m$ can be such that $m \in \{high, low\}$. These are modulated by synaptic gating (See Section \ref{synapticGating}). A small decay is also applied to active segments of neurons that did not become active.

\begin{equation}
\label{eqn:lastHTMeqn}
\triangle G^d_{ij} = p^{--} \dot{G}^d_{ij} \; where \; a^t_{ij} = 0 \; and \; \left\lVert\widetilde{G}_{ij}^d \circ A^{t-1}\right\rVert_1 > \theta
\end{equation}

The segments $G_{ij}$ are used to populate the weights $W^{DS}$ which are then used in Equation \ref{eqn:currentEquation}. 

We illustrate this higher-order sequence memory with an example. If a sequence $ABCD$ is presented for the first time, only proximal input will be activated by $A, B, C$ and $D$. All neurons in the corresponding columns will become active upon receiving the proximal input. Once the sequence is learned, when $A$ is active, $B$ will automatically be predicted. If the proximal input for $B$ is also received, only the neuron in the column $B$ which has been predicted will be active. If two sequences, $ABCD$ and $XCE$ are learned, then in the column representing $C$, one neuron gets activated when sequence $ABCD$ is presented, and a different neuron in the column gets activated when $XCE$ is presented. This is because although each neuron in the column represents pattern $C$, one neuron represents $C$ in the context of $ABCD$, and another neuron represents pattern $C$ in the context of $XCE$. For more details, please refer to \cite{Hawkins:2016}. 




\subsection{Novel Features Added to the Model}\label{addedFeas}

This section describes the features that we have added to the original HTM system \cite{Hawkins:2016}, as we described in Section \ref{addedFeas_bkg}. 

\begin{figure}[h!]
	\begin{center}
		\includegraphics[width=8cm]{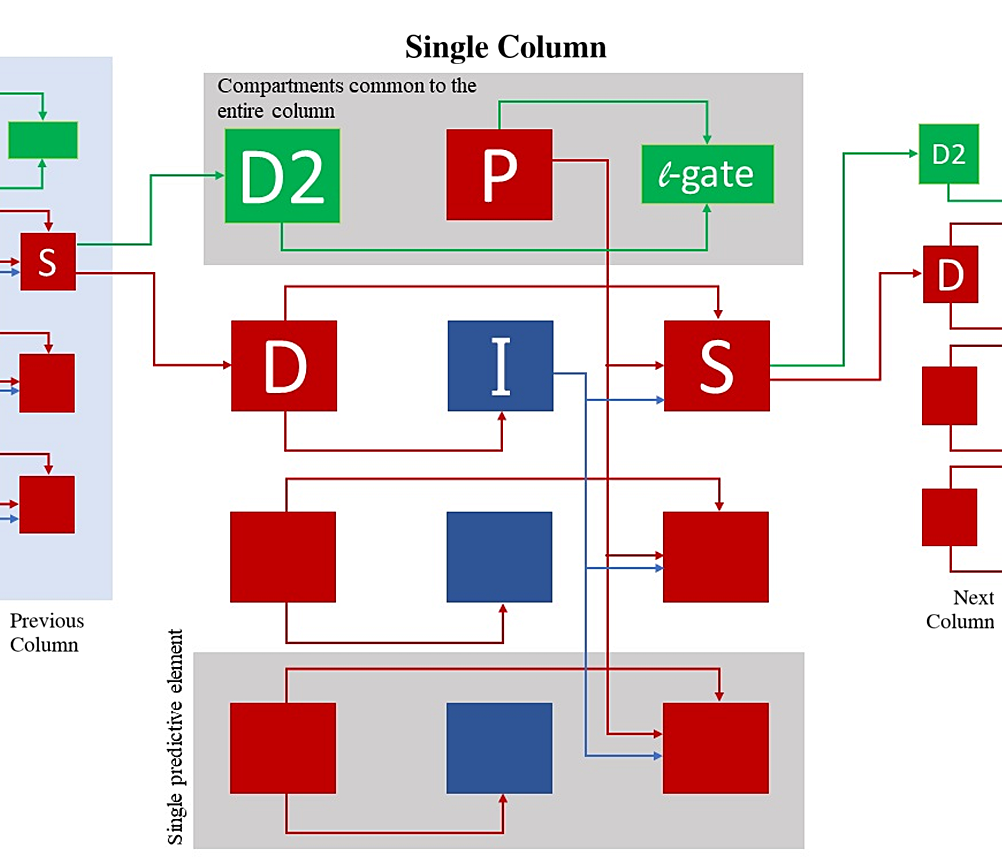}
	\end{center}
	\caption{\textbf{System Architecture}: The spiking version of HTM has been implemented with
 five LIF compartments, P- Proximal, D- Distal, I- Inhibition, S- Soma and D2- Distal 2
 compartments, and the l-gate. The P and D2 compartments and l gate are common to each
 column. Each neuron is represented by D, I and S compartments. P receives the proximal
 input and activates all S neurons in a column, all of which remain active in the absence
 of any distal input. D receives input from S neurons of other columns and is activated if
 any of its segments are active. D activates its corresponding I neuron and depolarizes its
 corresponding S neuron. In the same column, I inhibits all the S neurons other than its own. So
 if the D fires, all the other S neurons in the same column are inhibited. Thus, the HTM
 functionality is implemented. This is described in \cite{Billaudelle:2016}. In our paper, we have the following
 additions, denoted by green boxes and arrows. The S of column $x$ also activates the D2 of
 column $y$ if there is a LTM connection from $x$ to $y$. If both P and D2 are active during the
 same pattern presentation, then a correct prediction has been made due to LTM weights,
 and the l-gate becomes active. When the l-gate is active, the system learns with a higher
 learning rate, $p^+_{high}$ else the learning rate is lower ($p^+_{low}$).}\label{fig:spikingHTM}
\end{figure}

\begin{enumerate}
\item \textbf{Implementation of HTM as a Spiking Neural Network:} The HTM network has been implemented as a spiking neural network as shown in Figure \ref{fig:spikingHTM}. Each HTM neuron is modeled with three compartments, represented by three different LIF neurons - the compartment D or $distal$, the compartment I or $inhibition$, and the compartment S or $soma$. Each column has a P or \emph{head neuron} that collects the proximal input and activates all S neurons in the column.

Each compartment follows the Leaky Integrate and Fire (LIF) learning rule. The voltage $V_j^C$ of neuron $j$ in compartment $C$ is calculated as: 

\begin{eqnarray}
\tau_M^C\frac{dV_j^C}{dt} = -(V_j^C - V_{rest}) + RI_j^C(t) \\
V_j^C (t) > V_{th} \implies s_j^C = 1; V_j^C(t) = V_{reset}
\end{eqnarray}

where $\tau_M^C$ is the membrane time constant of compartment $C$. Compartment $C$ can refer to $D$, $I$, $S$, $D2$ or $L$ compartments. $s_j^C(t)$ is the spike train of neuron $j$ in the $C$ compartment and equals $1$ during the time neuron $j$ fires a spike and $0$ otherwise. After a spike is fired, the membrane voltage of the neuron is set to $V_{reset}$ and the neuron is refractory for the absolute refractory period $v_{ref}$, before it can fire again. 

The current $I_j^{CD}$ from compartment C to compartment D into each neuron $j$ is calculated as: 

\begin{eqnarray} 
\label{eqn:currentEquation}
I_j^{CD} = A^{CD} \sum_k W^{CD}_{jk} g^{CD}_k \\
\tau_g \frac{dg_k^{CD}}{dt} = -g_k^{CD}(t) \\
g_k^{CD}(t) = g_k^{CD}(t) + s_k^{CD}(t)
\end{eqnarray}

where $g_k^{CD}$ is the input kernel function. 

The compartments work as follows: 

\begin{enumerate}
\item The $P-compartment$ collects the proximal information and feeds forward to the $S-compartment$. This creates the current $I^{PS}$ from the $P-compartment$ to the $S-compartment$. 
\item The $D-compartment$ sums up the distal input from the previous time-step. (See next section for methodology).  If the $D-compartment$ is active, it feeds forward to the $I-compartment$ and $S-compartment$s respectively. It creates the currents $I^{DI}$ and $I^{DS}$ which is enough to activate the $I-compartment$, it can only depolarize the $S-compartment$. This is achieved because $A^{PS} >> A^{DS}$
\item The $I-compartment$ inhibits all $S-compartments$ other than the one corresponding to itself. The $S-compartment$ receives the input current as follows: 
\begin{equation}
I^S = I^{PS} + I^{DS} + I^{IS}
\end{equation}
\end{enumerate}

So the $S-compartment$ receives neurons from the $P-compartment$ for the proximal input, the $D-compartment$ for the distal input and the $I-compartment$ for the inhibition. The effect is - if there is no distal input, all $S-compartment$ neurons are active. Else, the $D-compartment$ depolarizes the corresponding $S-compartment$ and the $I-compartment$ inhibits all the other $S-compartments$ in the same column. This achieves the effect described in the previous section on the HTM functionality.  

These interactions are described in Figure \ref{fig:spikingHTM} and are based on\cite{Billaudelle:2016}. The only weights that are learned are the weights from the $D-compartment$ to the $S-compartment$, or the $W^{DS}$ weights, which are updated according to Equations \ref{eqn:firstHTMeqn} to \ref{eqn:lastHTMeqn}. 

\item \textbf{Inclusion of LTM weights:} The architecture, described in \cite{Billaudelle:2016} has been augmented with an additional compartment, D2 (colored in green, to denote an addition in our system) which incorporates the LTM weights. As in P, there is only one D2 neuron for each column. 

The $D2-compartment$ collects the proximal information and feeds forward to the $S-compartment$. This creates the current $I^{PS}$ from the $P-compartment$. The WAN known as SWOW-EN \cite{Deyne:2019} has been incorporated into the system as the LTM weights in the following manner. In SWOW-EN, R1 refers to the mode where the first word subjects gave when prompted with a cue word is considered and R123 refers to the mode where first three responses subjects gave is considered. A complete system diagram is given in Figure \ref{fig:spikingHTM}. The weights from the LTM to the D2 compartment are translated as follows. 

\begin{eqnarray}
\text{if} \;  W^{SWOW-EN}_{xy} \; = 1 \\
\forall j \in x', \forall k \in y' \; \; W^{D2S}_{jk} = 1
\end{eqnarray}

where $x$ and $y$ are items in the memory. $x'$ represents the set of all neurons in the columns corresponding to item $x$ in the $S$ and $y'$ corresponds to the neurons in all the columns in the $D2-compartment$ corresponding to item $y$. If any S neuron is activated in one column, it activates the D2 neurons connected to them in other columns. An activation of D2 represents a prediction due to the LTM weights. 

\item \textbf{Gating of Plasticity:} The following is the implementation of the gating of plasticity described in section \ref{addedFeas}. Gating of plasticity is achieved by the activation of the $l-gate$ which is a flag that modulates the learning rate, $p^+$. 
\label{synapticGating}

\begin{eqnarray}
if \; active_t(D2) \; \land \; active_t(P) \\ \nonumber 
l-gate = 1, \; p^+ = p^+_{high} \\ \nonumber
else \; l-gate = 0, \; p^+ = p^+_{low}
\end{eqnarray}

The $l-gate$ modulates the \emph{learning rate} in the system – if active, reinforcement occurs with a learning rate of $p^+_{high}$. Else, reinforcement occurs with the learning rate of $p^+_{low}$. The learning rates are used in equation \ref{eqn:learningEqn}.  

\item \textbf{Rehearsal:} This is incorporated in the system as follows. The system has two phases, the \emph{learning phase} and \emph{rehearsal phase}. During the \emph{learning phase}, the sequences are presented one after another, as is generally done in machine learning. In the \emph{rehearsal phase}, learning is different from the general machine learning practice. During this phase, a sequence is repeatedly iterated through. Once the last pattern of the sequence is presented, the sequence is repeated, i.e. the first pattern of the same sequence is again presented followed by the second and so on. During this phase, a \emph{p-c pair} which has been presented and whose permanence is less than the threshold, $\tau$ is repeated with a probability of $q$. A \emph{p-c pair} with a permanence of $\tau$ or more has $0$ probability of repetition.  In this manner, during the \emph{rehearsal phase}, a sequence is iterated through, with more attention given to unlearned patterns. After $i$ such iterations, due to fatigue, the system changes to another sequence.

\item \textbf{Constant Decay: } All weights are subject to constant decay, $\rho$ as described in Section \ref{addedFeas}. In the presence of constant decay, only weights that have been repeatedly reinforced stay active. Weights that are a result of noise decay to 0. 
\end{enumerate}

\section{Simulations and Results}

In the first simulation, we demonstrate basic HTM functionality. The other experiments demonstrate learning in the presence of long-term memory. As described earlier in Section \ref{addedFeas}, LTM refers to the system initialized with WAN weights. 

\subsection{Original HTM Functionality}\label{sec:OriginalHTM}


\begin{figure}[h!]
	\begin{center}
		\includegraphics[width=8cm]{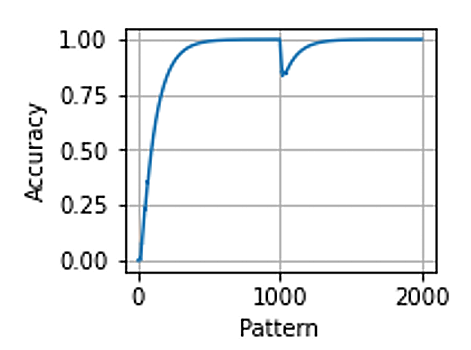}
	\end{center}
	\caption{\textbf{Original HTM Functionality:} This figure demonstrates the original HTM functionality. It shows higher-order continual online learning and is equivalent to \cite{Hawkins:2016}, Figure
 6. After 1000 sequence presentations, the entire dataset is changed and the system learns a
 changed dataset.}\label{fig:originalHTM}
\end{figure}

This simulation was done with the basic HTM system, without adding the functionality described in Section \ref{addedFeas}, such as \emph{D2 compartment} and \emph{l-gate}. This is to show that our system is indeed based on HTM and is able to reproduce its functionality. The experiment in this section is similar to Experiment 4.1 in \cite{Cui:2016}, with some differences. In the \textit{original experiment}, the input is
a stream of elements, where every element is converted to a
2\% sparse activation of mini-columns. This network learns a predictive model of the data based on the transitions observed in the input stream.  In our system, we introduce some differences. First, it was not clear how many sequences were used in \cite{Cui:2016} - the rules were stipulated, but the number of sequences was not. We trained on eight sequences. Secondly, we use a start pattern at the beginning of each sequence, which simplifies the problem. Humans generally know the start of a sequence, as they are aware of the context in which they are learning. 

In this experiment, the system learns a continuous stream of sequences interspersed with noise. There are eight sequences as given in Table \ref{tab:dataset}. 

\begin{table}
  \begin{center}
	\begin{tabular}{ |c|c| } 
		\hline
		XABCDE & KPQRIJ \\ 
		YABCFG & KPQRLM \\ 
		XMNODE & SPQRUV \\
		YMNOFG & SPQRWX \\ 
		\hline		
    \end{tabular} 
  \end{center}
	\caption{\textbf{Dataset used for testing original HTM functionality:} This dataset has a stream of letters to be learnt by the HTM model. The dataset is created using rules stipulated in \cite{Cui:2016}, Figure 3.} \label{tab:dataset}
\end{table}

These have been created with rules described in  \cite{Cui:2016}, Figure 3. An epoch consists of the presentation of one of the eight sequences, and noise.  Accuracy is calculated after the fourth element of the sequence is presented. The equation used for calculating accuracy is as follows: 

\begin{eqnarray} \label{eqn:calcAcc}
P(\text{Column is Predicted}) | P(\text{Column is Active}) = 
\\ \frac{\text{No. Columns Active} \cap \text{No. Columns Predicted}}{\text{No. Columns Active}} 
\end{eqnarray}

If all columns that are active have been predicted, this accuracy will be $1$, otherwise it will reduce proportionately. After $1000$ sequence presentations, all the sequences are modified. The system has to re-learn the new sequences. 

We expect a maximum accuracy of $100\%$, which should be achieved once the system has learned. The results are in Figure \ref{fig:originalHTM}. 
As we can see the learning is similar to the one in \cite{Cui:2016}. 

\subsection{Comparing LTM Weights and Random Weights} \label{sec:LTMrandom}

\begin{figure}[h!]
	\begin{center}
		\includegraphics[width=8cm]{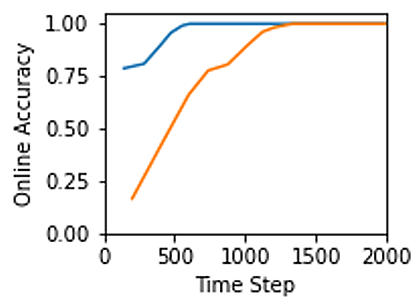}
	\end{center}
	\caption{\textbf{Comparing LTM and Random Weights:} The blue color shows learning in the system initialized with LTM weights, while the orange color shows learning in the system with random weights. Here, we show only the \emph{rehearsal} phase.}\label{fig:ltmRandom}
\end{figure}

In the rest of the simulations, the system has all the functionality that has been described in this paper. (See Section \ref{addedFeas}). We trained the system on five excerpts from well-known nursery rhymes given in Table \ref{tab:datasetPoems}.   

\begin{table}
\begin{center}
	\begin{tabular}{ |l|l| } 
		\hline
		1. & Wheels of the bus go round and round \\
		&  Round and round, round and round \\ 
		\hline
		2. & One, two, buckle my shoe  \\
		& Three, four, shut the door \\ 
		\hline
		3. & Way up in the sky, \\
		& The little birds fly \\
		& When done in the nest \\
		& The little birds rest \\ 
		\hline	
		4. & Head and shoulders, knees and toes, knees and toes \\
		& And eyes and ears and mouth and nose \\ 
		\hline
		5. & Cows give milk that we can drink  \\
		& To make us grow up strong \\ 
		\hline
	\end{tabular}
\end{center}
\caption{\textbf{Dataset used for Experiments \ref{sec:LTMrandom} - \ref{sec:discrim}:} This contains excerpts from well-known English nursery rhymes.} \label{tab:datasetPoems}
\end{table}

We performed stop-word removal on the system as follows. Any word that was present in the set of 100 most frequent words in English \cite{Oxford:2006} was removed. 

In these experiments, the system is initialized with (1) the LTM weights and (2) random weights which are created as follows. Let us say that the number of connections in the LTM weights is $m$. A zero connectivity matrix is initialized, and $m$ random connections are added. Thus, the system with random weights has the same number of connections and dimensionality as the system with LTM weights. 

The system had a constant decay, $\rho = 1 \times 10^{-7}$s. It went through a \emph{learning phase} for $5$ epochs, and \emph{rehearsal phase} for $20$ epochs, where patterns were shuffled and presented to the system. We show only the \emph{rehearsal phase} because \emph{learning phase} had only a few epochs. During the \emph{learning phase}, connections that were predicted by the LTM weights were quickly learned, and others were not adequately learned. The starting accuracy during the \emph{rehearsal phase} was the accuracy obtained during the \emph{learning phase}. Therefore, we show the \emph{rehearsal phase} alone. 

As can be seen learning occurs much faster in the system with LTM weights compared to the system with random weights. In the next experiment, we examine this learning in the presence of noise. 

\subsection{Comparing LTM Weights and Random Weights with Noise} \label{sec:LTMrandomNoise}

\begin{figure}[h!]
	\begin{center}
		\includegraphics[width=8cm]{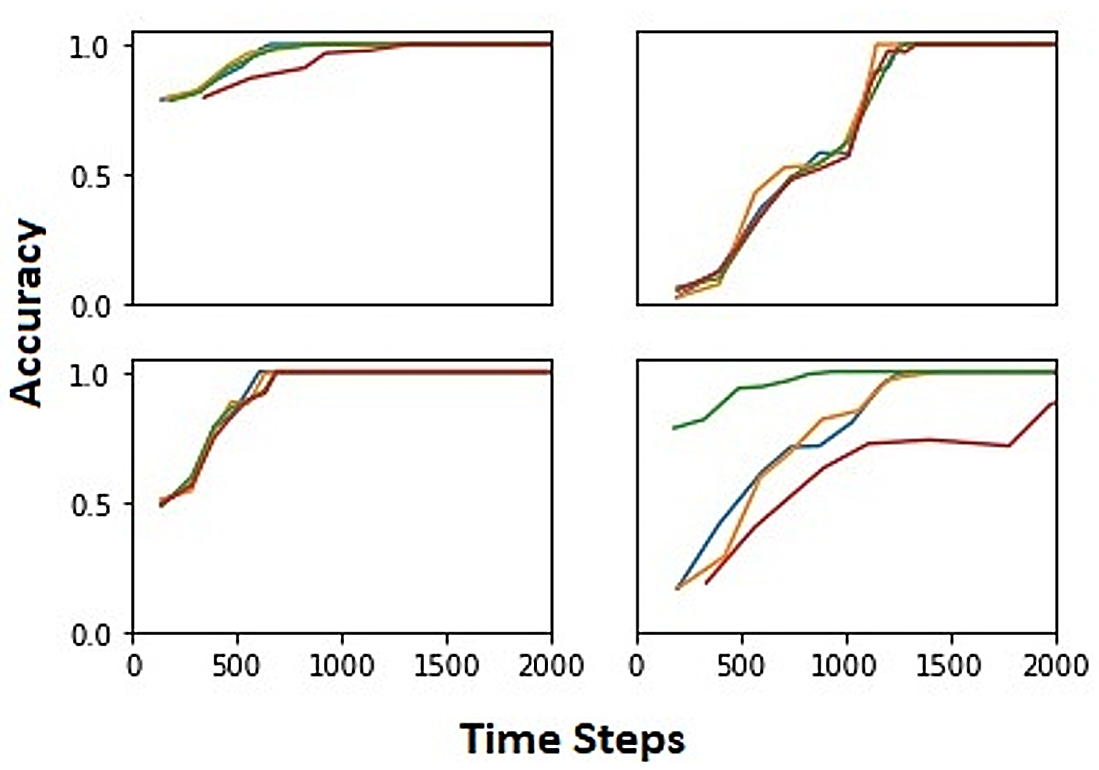}
	\end{center}
	\caption{\textbf{Comparing LTM and Random Weights with Noise:} This is the comparison
 of learning in the system initialized with LTM weights (left 2 images) and random weights
 (right 2 images) with noise. Each of the plots feature 4 separate values of noise, green - 10\% noise, orange - 30\%, blue - 50\% noise and red - 70\% noise. The top $2$ images have a constant
 decay $\rho = 1 \times 10^{-7}$ and the bottom two images have a constant decay $\rho = 3 \times 10^{-7}$.}\label{fig:ltmRandomNoise}
\end{figure}

This experiment is identical to the previous experiment, except that we added noise. Noise sequences are between $7-13$ random words long. This is equivalent to the range of the length of meaningful sequences. Each random word is represented by set of $6$ columns selected at random. This set is not necessarily identical to that of any existing word. 

During training and rehearsal, there is a probability of $n$ that the system encounters a noise sequence, and that of $1-n$ that it encounters a sequence in the dataset. We tested on $4$ values of $n$, $0.1$, $0.3$, $0.5$ and $0.7$. 

As before, we compare LTM and random weights. Two decay ($\rho$) terms were compared, a decay of $1 \times 10^{-7}$s and $3 \times 10^{-7}$s. The results are in Figure \ref{fig:ltmRandomNoise}. As can be seen in the results, continuous learning has to overcome the effects of noise and decay of weights. Learning in the presence of LTM weights occurs very quickly, both with and without noise. With random weights, noise has less effect on accuracy when decay levels are low. When decay levels are higher the effects of noise are more pronounced. The effects of constant learning have to overcome noise and decay. In the system with random weights, during the \emph{learning phase}, very little learning has taken place. Many connections between words in a sequence are unlearned at the beginning of the \emph{rehearsal phase}, and accuracy is close to $0$. With many repetitions of unlearned items, there is some increase in accuracy. However, since there are ordered sequences of many words, and all of them require learning for further increase in accuracy, the system requires many repetitions.  Once there are only a few connections to be learned, due to rehearsal and repetition, learning focuses on unlearned items which can then quickly learn.   

\subsection{Discriminative Learning} \label{sec:discrim}

\begin{figure}[h!]
	\begin{center}
		\includegraphics[width=8cm]{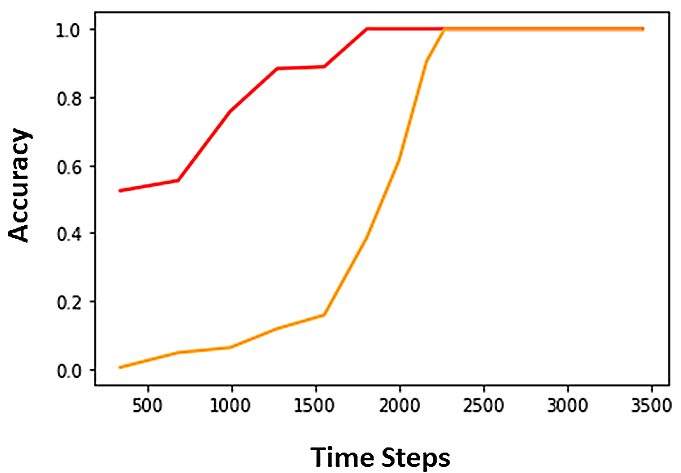}
	\end{center}
	\caption{This shows the learning of the sequences in two scenarios- red- with meaningful
 sequences from the poems dataset, and orange- with a random sequence of words.}\label{fig:discriminativeLearning}
\end{figure}

In this experiment, the system was initialized with LTM weights. The system was trained on (1) The regular dataset, which is the excerpts from poems, and (2) random groups of words (nonsense sequences). The sequences in (2) were generated by selecting a random ordered set of words from lengths between $6-12$. The \emph{starter} pattern was appended to the beginning of each of these sequences. We examined the time taken to learn both these. The results are presented in Figure \ref{fig:discriminativeLearning}. 

As can be seen, in the presence of the LTM weights, the system is able to discriminate meaningful songs from random groups of words. Excerpts from poems are learned very quickly in the presence of the LTM. While it is still possible to learn nonsense sequences, it takes a lot more time and repetitive training to learn these.  This shows that few-shot sequence learning is possible for the system but only for sequences that make sense in the context of the language.

An alternative way to look at the results is to see the faster learning of sensible sequences as a beneficial cognitive feature that keeps the system from learning nonsense. The modulation of the learning rate by prediction is can thus be seen as an adaptive filtering mechanism.

\section{Conclusion}
\label{discussion}

It is important to note that work in this paper is only a first step to the exploration of meaning. As seen in the background section, WANs are one of the best representations of the mental organization of words. However, WAN combines and flattens out different kinds of word associations including homonymy, synonymy, word co-occurrence and so on. This is beneficial as a first step exploration, as it can embody many kinds of word relationships with just one set of weights. The downside is that it is not possible to tease apart the different semantic influences that shape up the meaning of words. 

As we propose in the MCM framework, meaning is influenced not just by association between words. Here we have implemented a small part of the entire MCM framework. A computational implementation of the entire MCM would include the influence of contexts, features, and associations between words. Different kinds of questions on learning, inference and memory search can be explored with such a framework.   



\balance
\bibliographystyle{unsrt}
\bibliography{biblio}

\begin{thebibliography}{10}

\bibitem{Hauk:2011}
O.~Hauk and F.~Pulvermuller.
\newblock The lateralization of motor cortex activation to action-words.
\newblock {\em Frontiers in Human Neuroscience}, 2011.

\bibitem{Carota:2017}
F.~Carota, N.~Kriegeskorte, H.~Nili, and F.~Pulvermuller.
\newblock Representational similarity mapping of distributional semantics in left inferior frontal, middle temporal, and motor cortex.
\newblock {\em Cerebral Cortex}, 27:294--309, 2017.

\bibitem{Barros:2011}
A.~Barros-Loscertales, J.~Gonzalez, F.~Pulvermuller, Ventura-Campos N., J.C. Bustamante, V.~Costumero, and C.~Avila.
\newblock Reading salt activates gustatory brain regions: f{MRI} evidence for semantic grounding in a novel sensory modality.
\newblock {\em Cerebral Cortex}, 22:2554--2563, 2011.

\bibitem{Barsalou:1999}
L.W. Barsalou.
\newblock Perceptual symbol systems.
\newblock {\em Behavioral and Brain Sciences}, 1999.

\bibitem{Barsalou:2003}
L.W. Barsalou.
\newblock Abstraction in perceptual symbol systems.
\newblock {\em Philosophical Transactions of the Royal Society of London Series B: Biological Sciences}, 358:1177--1187, 2003.

\bibitem{Barsalou:2008}
L.W. Barsalou.
\newblock Grounded cognition.
\newblock {\em Annual Reviews of Psychology}, 59:617--645, 2008.

\bibitem{Tomasello:2018}
R.~Tomasello, M.~Garagnani, T.~Wennekers, and F.~Pulvermuller.
\newblock A neurobiologically constrained cortex model of semantic grounding with spiking neurons and brain-like connectivity.
\newblock {\em Frontiers in Computational Neuroscience}, 2018.

\bibitem{Whitney:2011}
C.~Whitney, M.~Kirk, J.~O'Sullivan, M.A. Lambon-Ralph, and E.~Jefferies.
\newblock The neural organization of semantic control: {TMS} evidence for a distributed network in left inferior frontal and posterior middle temporal gyrus.
\newblock {\em Cerebral Cortex}, 21:1066--1075, 2011.

\bibitem{Iyer:2009}
L.R. Iyer, S.~Doboli, A.~A. Minai, V.~R. Brown, D.~S. Levine, and P.~B. Paulus.
\newblock Neural dynamics of idea generation and the e ects of priming.
\newblock {\em Neural Networks}, 22:674--686, 2009.

\bibitem{Iyer:2021}
L.R. Iyer and A.A. Minai.
\newblock {\em Creativity and Innovation: Cognitive, Social and Computational Approaches}, chapter Models of creativity and ideation: An overview, pages 21--45.
\newblock Springer, Switzerland, 2021.

\bibitem{Minai:2021a}
A.A. Minai, L.R. Iyer, and S.~Doumit.
\newblock {\em Creativity and Innovation: Cognitive, Social and Computational Approaches}, chapter {IDEA} itinerant dynamics with emergent attractors: A neural model for conceptual combination, pages 195--227.
\newblock Springer, Switzerland, 2021.

\bibitem{Iyer:2010}
L.R. Iyer, V.~Venkatesan, and A.A. Minai.
\newblock Neurocognitive spotlights: Configuring domains for ideation.
\newblock In {\em The 2010 International Joint Conference on Neural Networks (IJCNN)}, pages 1--8, 2010.

\bibitem{Minai:2021b}
A.A. Minai, S.~Doboli, and L.R. Iyer.
\newblock {\em Models of creativity and ideation: An overview}, chapter Creativity and Innovation: Cognitive, Social and Computational Approaches, pages 21--45.
\newblock Springer, Switzerland, 2021.

\bibitem{Polyn:2009}
S.M. Polyn, K.A. Norman, and M.J Kahana.
\newblock A context maintenance and retrieval model of organizational processes in free recall.
\newblock {\em Psychological Review}, pages 129--156, 2009.

\bibitem{Nelson:2004}
D.L. Nelson, C.L. Mcevoy, and T.A. Schreiber.
\newblock The university of south florida free association, rhyme, and word fragment norms.
\newblock {\em Behaviour Research Methods, Instruments \& Computers}, 36:402--407, 2004.

\bibitem{Deyne:2019}
S.D. Deyne, D.J. Navarro, A.~Perfors, M.~Brysbaert, and G.~Storms.
\newblock The small world of words english word association norms for over 12,000 cue words.
\newblock {\em Behavior Research Methods}, 2019.

\bibitem{Otgaar:2012}
H.~Otgaar, M.~Peters, and M.~L. Howe.
\newblock Dividing attention lowers children's but increases adults' false memories.
\newblock {\em Journal of Experimental Psychology: Learning, Memory, and Cognition}, 38:204--210, 2012.

\bibitem{Gough:1976}
H.G. Gough.
\newblock Studying creativity by means of word association tests.
\newblock {\em Journal of Applied Psychology}, 61:348--353, 1976.

\bibitem{Benedek:2012}
M.~Benedek, T.~Könen, and A.~C. Neubauer.
\newblock Associative abilities underlying creativity.
\newblock {\em Psychology of Aesthetics, Creativity, and the Arts}, 6:273--281, 2012.

\bibitem{Kenett:2019}
Kenett Y.N. and M.~Faust.
\newblock A semantic network cartography of the creative mind.
\newblock {\em Trends in Cognitive Sciences}, 2019.

\bibitem{Kenett:2018}
Kenett Y.N., O.~Levy, D.Y. Kenett, H.E. Stanley, M.~Faust, and S.~Havlin.
\newblock Flexibility of thought in high creative individuals represented by percolation analysis.
\newblock {\em PNAS}, 2018.

\bibitem{Ronald:1964}
Ronald~D. W.
\newblock Are normal word association norms suitable for schizophrenics?
\newblock {\em Psychological Reports}, 14:121--122, 1964.

\bibitem{Brendan:2005}
Brendan A.M., Theo C.M., Jakob L., and Steven C.
\newblock Quantitative assessment of the frequency of normal associations in the utterances of schizophrenia patients and healthy controls.
\newblock {\em Schizophrenia Research}, 78:219--224, 2005.

\bibitem{Manschreck:2012}
T.C. Manschreck, A.M. Merrill, G.~Jabbar, J.~Chun, and L.E. DeLisi.
\newblock Frequency of normative word associations in the speech of individuals at familial high-risk for schizophrenia.
\newblock {\em Schizophrenia Research}, 40:99--103, 2012.

\bibitem{Stamps:1995}
J~Stamps.
\newblock Motor learning and the value of familiar space.
\newblock {\em The American Naturalist}, pages 41--58, 1995.

\bibitem{Camina:2017}
E~Camina and F~Guell.
\newblock The neuroanatomical, neurophysiological and psychological basis of memory: Current models and their origins.
\newblock {\em Frontiers in Pharmacology}, 8, 2017.

\bibitem{Cui:2016}
Y.~Cui, S.~Ahmad, and J.~Hawkins.
\newblock Continuous online sequence learning with an unsupervised neural network model.
\newblock {\em Neural Computation}, 2016.

\bibitem{Hawkins:2016}
J.~Hawkins and S.~Ahmad.
\newblock Why neurons have thousands of synapses, a theory of sequence memory in neocortex.
\newblock {\em Frontiers in Neural Circuits}, 10, 2016.

\bibitem{Hawkins:2017}
J.~Hawkins, S.~Ahmad, and Y.~Cui.
\newblock A theory of how columns in the neocortex enable learning the structure of the world.
\newblock {\em Frontiers in Neural Circuits}, 11, 2017.

\bibitem{Basu:2016}
J.~Basu, J.D. Zaremba, S.K. Cheung, F.L. Hitti, B.V. Zemelman, A.~Losonczy, and S.A. Siegelbaum.
\newblock Gating of hippocampal activity, plasticity, and memory by entorhinal cortex long-range inhibition.
\newblock {\em Science}, 351, 2016.

\bibitem{Lehmann:2015}
M.~Lehmann.
\newblock Rehearsal development and development of iterative recall processes.
\newblock {\em Frontiers in Psychology}, 6, 2015.

\bibitem{Parle:2006}
M.~Parle, N.~Singh, and M.~Vasudevan.
\newblock Regular rehearsal helps in consolidation of long term memory.
\newblock {\em Journal of Sports Science and Medicine}, pages 80--88, 2006.

\bibitem{Bonstrup:2019}
M.~Bonstrup, I.~Iturrate, R.~Thompson, G.~Cruciani, N.~Censor, and L.G. Cohen.
\newblock A rapid form of online consolidation in skill learning.
\newblock {\em Current Biology}, pages 1346--1351, 2019.

\bibitem{Sievers:1944}
D.W. Sievers.
\newblock The play rehearsal schedule and its psychology.
\newblock {\em Quarterly Journal of Speech}, 30:80--84, 1944.

\bibitem{Ackermann:2017}
B.J. Ackermann.
\newblock How much training is too much?
\newblock {\em Medical Problems of Performing Artist}, pages 61--62, 2017.

\bibitem{Zucker:2002}
R.S. Zucker and W.G. Regehr.
\newblock Short term synaptic plasticity.
\newblock {\em Annual Reviews of Physiology}, 64:355--405, 2002.

\bibitem{Leimer:2019}
P.~Leimer, M.~Herzog, and W.~Senn.
\newblock Synaptic weight decay with selec tive consolidation enables fast learning without catastrophic forgetting.
\newblock {\em bioRxiv}, 2019.

\bibitem{Panda:2018}
P.~Panda, J.M. Allred, S.~Ramanathan, and K.~Roy.
\newblock {ASP}: Learning to forget with adaptive synaptic plasticity in spiking neural networks.
\newblock {\em {IEEE} Journal on Emerging and Selected Topics in Circuits and Systems}, 8:51--64, 2018.

\bibitem{Tetzlaff:2013}
C.~Tetzlaff, C.~Kolodziejski, M.~Timme, M.~Tsodyks, and F.~Worgotter.
\newblock Synaptic scaling enables dynamically distinct short- and long-term memory formation.
\newblock {\em {PLOS} Computational Biology}, 2013.

\bibitem{Billaudelle:2016}
S.~Billaudelle and S.~Ahmad.
\newblock Porting {HTM} models to the heidelberg neuromorphic computing platform.
\newblock {\em arXiv}, 2016.

\bibitem{Oxford:2006}
{\em The Oxford English Corpus}.
\newblock Oxford, 2006.

\end{thebibliography}

\end{document}